\documentclass[journal]{IEEEtran}
\usepackage{cite}

\ifCLASSINFOpdf
\else
\fi
\usepackage{amsmath}
\usepackage{multirow}
\usepackage{graphicx}
\usepackage[export]{adjustbox}

\newcommand{\specialcell}[2][c]{%
  \begin{tabular}[#1]{@{}c@{}}#2\end{tabular}}

\usepackage[table]{xcolor}
\usepackage{xcolor}

\begin{document}
%
\title{DeepSleepNet-Lite: A Simplified Automatic Sleep Stage Scoring Model with Uncertainty Estimates}
%
%
%

\author{Luigi~Fiorillo,~\IEEEmembership{Member,~IEEE,}
        Paolo~Favaro,~\IEEEmembership{Member,~IEEE,}
        and~Francesca~Dalia~Faraci,~\IEEEmembership{Member,~IEEE}
\thanks{L. Fiorillo is with the Institute of Informatics, University of Bern, Bern, Switzerland, and the Institute for Information Systems and Networking, University of Applied Sciences and Arts of Southern Switzerland, Lugano, Switzerland, e-mail: luigi.fiorillo@supsi.ch.}
\thanks{P. Favaro is with the Institute of Informatics, University of Bern, Bern, Switzerland.}
\thanks{F. D. Faraci is with the Institute for Information Systems and Networking, University of Applied Sciences and Arts of Southern Switzerland, Lugano, Switzerland.}
\thanks{Manuscript received January 15, 2021; revised XX XX, 2021}}

%
%

\markboth{Journal of \LaTeX\ Class Files,~Vol.~XX, No.~XX, XX~2021}%
{Shell \MakeLowercase{\textit{et al.}}: Bare Demo of IEEEtran.cls for IEEE Journals}
%



\maketitle

\begin{abstract}
Deep learning is widely used in the most recent automatic sleep scoring algorithms. Its popularity stems from its excellent performance and from its ability to directly process raw signals and to learn feature from the data. Most of the existing scoring algorithms exploit very computationally demanding architectures, due to their high number of training parameters, and process lengthy time sequences in input (up to 12 minutes). Only few of these architectures provide an estimate of the model uncertainty.
In this study we propose DeepSleepNet-Lite, a simplified and lightweight scoring architecture, processing only 90-seconds EEG input sequences. We exploit, for the first time in sleep scoring, the \textit{Monte Carlo dropout} technique to enhance the performance of the architecture and to also detect the uncertain instances.
The evaluation is performed on a single-channel EEG Fpz-Cz from the open source Sleep-EDF expanded database.
DeepSleepNet-Lite achieves slightly lower performance, if not on par, compared to the existing state-of-the-art architectures, in overall accuracy, macro F1-score and Cohen's kappa (on Sleep-EDF v1-2013 $\pm$30mins: 84.0\%, 78.0\%, 0.78; on Sleep-EDF v2-2018 $\pm$30mins: 80.3\%, 75.2\%, 0.73). \textit{Monte Carlo dropout} enables the estimate of the uncertain predictions. By rejecting the uncertain instances, the model achieves higher performance on both versions of the database (on Sleep-EDF v1-2013 $\pm$30mins: 86.1.0\%, 79.6\%, 0.81; on Sleep-EDF v2-2018 $\pm$30mins: 82.3\%, 76.7\%, 0.76).
Our lighter sleep scoring approach paves the way to the application of scoring algorithms for sleep analysis in real-time.
\end{abstract}


%
\IEEEpeerreviewmaketitle

\section{Introduction}
%
%
%
%
\IEEEPARstart{G}{ood} sleep plays a crucial role in human well-being, and sleep disorders represent a significant and an increasing public health problem \cite{ohayon2011}. Polysomnography (PSG) is used in sleep medicine as a diagnostic tool, so as to objectively analyze the quality of sleep and the common sleep pathologies - e.g. sleep breathing disorders, narcolepsy, sleep-related movement disorders \cite{penzel2000}. 
Electroencephalography (EEG), electrooculography (EOG), electromyography (EMG) and electrocardiography (ECG) signals are  essential for the PSG exam. 
The physicians extract sleep cycle information through the well-known sleep stage scoring procedure, according to the AASM guidelines \cite{iber2007aasm}. The whole-night sleep recording is divided into 30-second windows, called epochs, and each epoch is classified into one of the following five sleep stages: wakefulness W, stage N1, stage N2, stage N3, and stage R (REM sleep). 
This manual sleep stage classification is obviously time-consuming and is affected by human error - several works report high values of inter- and intra-scorer variabililty \cite{berthomier2020}. Since 1960, a wide variety of techniques have been devised in an effort to automate this procedure. Still, up to now, no system has completely replaced the physician. \\
In the last decades, deep learning algorithms have been widely used to solve the sleep scoring task automatically. A thorough review of the
application of deep learning architectures to sleep scoring can be found in \cite{fiorillo2019automated}. Autoencoders \cite{tsinalis2016automatic}, deep neural networks (DNNs) \cite{dong2018mixed}, convolutional neural networks (CNNs) \cite{tsinalis2016automaticCNN, vilamala2017deep, chambon2018deep, cui2018automatic, olesen2018deep, patanaik2018end, sors2018convolutional, yildirim2019deep}, recurrent neural networks (RNNs) \cite{michielli2019cascaded} and several combination of them \cite{supratak2017deepsleepnet, biswal2018, malafeev2018automatic, stephansen2018, back2019, mousavi2019, phan2020, supratak2020, phan2021} have been recently proposed. The main advantage of all these networks is the ability to learn features directly from raw data, by taking into account the temporal dependency among the sleep stages. However, the architectures of these models are quite complex, a high number of parameters need to be trained. The most recent ones process lengthy time sequences in input - i.e. up to 12 minutes - using RNNs, thus requiring extra resources to buffer the PSG input and making them unsuitable in home-monitoring and in real-time applications. As a rule, deep architectures with a high number of layers and parameters need to be trained on large databases to prevent overfitting. In different scenarios sleep datasets have a limited number of labeled PSG samples available. Lighter architectures may be better suited if the model needs to be trained from scratch.
We found only two architectures \cite{patanaik2018end,stephansen2018} performing the automatic sleep scoring and also providing an estimate of the model uncertainty. In \cite{patanaik2018end} they use an additional classification block-2 (i.e. multilayer perceptron in cascade to the deep convolutional scoring architecture) to output the final sleep stage and the associated relative confidence score. In contrast, \cite{stephansen2018} trains 16 different models and uses the relative model variance to estimate the uncertain predictions. It is important to know the level of confidence of each prediction, as it could be the key to identify the misclassified sleep stages.

In this paper, we propose DeepSleepNet-Lite, a simplified and lightweight automatic sleep scoring architecture. It provides the predicted sleep stages along with an estimate of their uncertainty. The major advantage is that it does not require any additional computation over the baseline architecture to provide the estimate.\\
The two main contributions of this paper are: 1) the optimization of a simple feed-forward sleep scoring architecture, that processes only 90-second single-channel EEG in input; 2) the application of the \textit{Monte Carlo dropout} sampling technique, using \textit{dropout} at test time to capture the model uncertainty and to enhance the performance of the scoring system.
In Section \ref{architecture} we describe the architecture, the training algorithm and the regularization techniques used in our scoring system. In Section \ref{calibration} we briefly present the label smoothing technique used to calibrate the scoring architecture. Moreover we propose a new conditional probability distribution computed over the targets (i.e. our prior knowledge), and used to smooth our labels. In Section \ref{uncertainty} we present the \textit{Monte Carlo dropout} sampling technique, and its application within our sleep scoring system to estimate the uncertainty of the model. In the last sections, we demonstrate the efficiency of label smoothing and \textit{Monte Carlo dropout} techniques in both calibrating and enhancing the performance of our model. We also demonstrate the efficiency of the uncertainty estimate procedure, by showing that it is able to identify the most challenging sleep stage predictions. We finally show that DeepSleepNet-Lite achieves performance on par with most up-to-date scoring systems.


\section{DeepSleepNet-Lite}
\label{architecture}

The architecture of DeepSleepNet-Lite is strongly inspired by \textit{DeepSleepNet} from Supratak \cite{supratak2017deepsleepnet}. Unlike the original network, we have employed only the first \textit{representation learning} part, and trained it with a \textit{sequence-to-epoch} learning approach. The architecture receives in input a sequence of PSG epochs, and it predicts the corresponding target of the central epoch of the sequence. In \cite{fiorillo2020EMBC} we had already shown that the first \textit{representation learning} part of the architecture, trained with a small temporal context - 90-second epochs, does most of the work on a small-sized database.

\subsection{The Architecture}

\begin{figure}[t]
  \includegraphics[width=\linewidth]{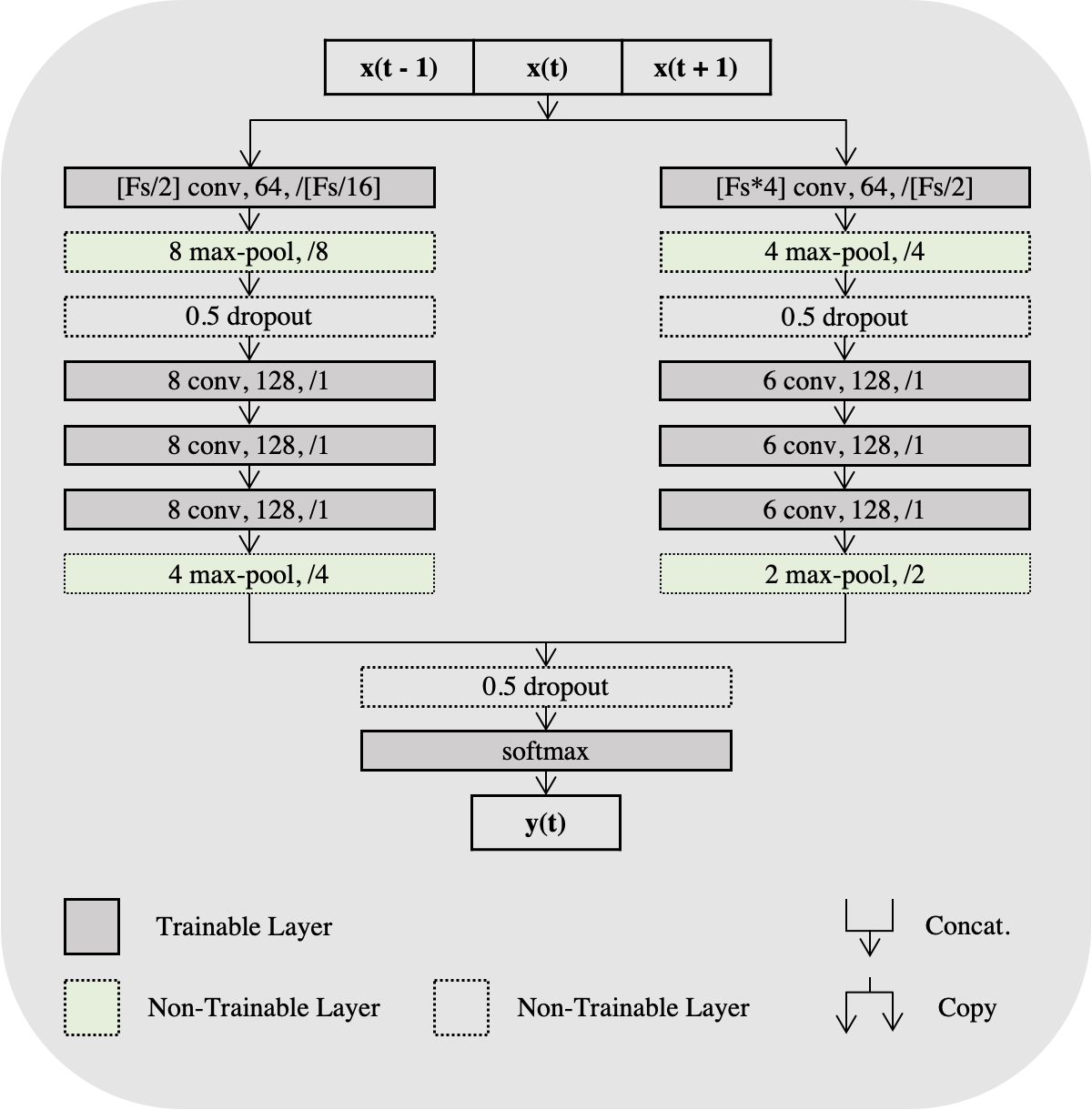}
  \caption{An overview of the \textit{representation learning} architecture from \cite{supratak2017deepsleepnet}, with our \textit{sequence-to-epoch} input-output training approach.}
  \label{fig:model}
\end{figure}

The \textit{representation learning} architecture consists of two parallel CNNs employing small ($CNN\textsubscript{$\theta$\textsubscript{\textit{S}}}$) and large ($CNN\textsubscript{$\theta$\textsubscript{\textit{L}}}$) filters at the first layer. The small filter has been used to extract high-time resolution patterns, while the large filter has been used to extract high-frequency resolution patterns. The idea behind the use of the small and large filter sizes comes from the way the signal processing experts define the trade-off between temporal and frequency precision in the feature extraction procedure \cite{cohen2014}.
Each CNN section consists of four convolutional layers and two max-pooling layers. Each convolutional layer executes three operations: a one-dimensional convolution of the filters with the 90-second epochs, a batch normalization \cite{ioffe2015} and an element-wise rectified linear unit (ReLU) activation function. The filter size, the number of filters and the stride size of each \textit{conv} layer are defined in Fig.~\ref{fig:model}. The pooling layer is used to downsample the input. In each \textit{max-pool} unit the pooling size and the stride size are specified.\\

The 90-second EEG signal \textbf{x}\textsubscript{i} is given in input to the convolutional neural networks $CNN\textsubscript{$\theta$\textsubscript{\textit{S}}}$ and $CNN\textsubscript{$\theta$\textsubscript{\textit{L}}}$. The parameters $\theta$ of each convolutional neural network are independently trained, so as to return in output two feature vectors \textbf{h}\textsubscript{i}\textsuperscript{\textit{S}} and \textbf{h}\textsubscript{i}\textsuperscript{\textit{L}}. The outputs are concatenated in \textbf{f\textsubscript{i}}, then forwarded to the \textit{softmax} layer. \\

\begin{equation}
\textbf{h}\textsubscript{i}\textsuperscript{\textit{S}} = CNN\textsubscript{$\theta$\textsubscript{\textit{S}}}(\textbf{x}\textsubscript{i}) \label{eq_net1}
\end{equation}
\begin{equation}
\textbf{h}\textsubscript{i}\textsuperscript{\textit{L}} =  CNN\textsubscript{$\theta$\textsubscript{\textit{L}}} (\textbf{x}\textsubscript{i})\label{eq_net2}
\end{equation}
\begin{equation}
\textbf{f}\textsubscript{i} = \textbf{h}\textsubscript{i}\textsuperscript{\textit{S}} \vert \vert \textbf{h}\textsubscript{i}\textsuperscript{\textit{L}} \label{eq_net3}
\end{equation}

The softmax function, together with the cross-entropy loss function, is used to train the model to output the logits \textbf{z}\textsubscript{i} and the probability for the five mutually exclusive classes.

\begin{equation}
\textbf{z}\textsubscript{i} = \textbf{W}\textsuperscript{T}\textbf{f}\textsubscript{i} + \textbf{b} \label{eq_net4}
\end{equation}
\begin{equation}
\hat{p}\textsubscript{i,k} = \frac{exp(z\textsubscript{i,k})}{\Sigma\textsubscript{j}exp(z\textsubscript{i,j})} \label{eq_net5}
\end{equation}

where \textit{$\theta$ = \{\textbf{W},\textbf{b}\}} are the parameters of the softmax layer, $j$ is the index of the vector \textbf{z}, $\hat{p}\textsubscript{i,k}$ is the output probability of class $k$ associated to x(t), the centred 30-second signal in \textbf{x}\textsubscript{i}.\\

All the model specifications are reported in Fig.~\ref{fig:model}, equally to the first \textit{representation learning} in \cite{supratak2017deepsleepnet}.

\subsection{Training Algorithm}

\begin{figure}[h!]
  \adjincludegraphics[width=\linewidth, trim={{.02\width} 0 {.04\width} 0},clip]{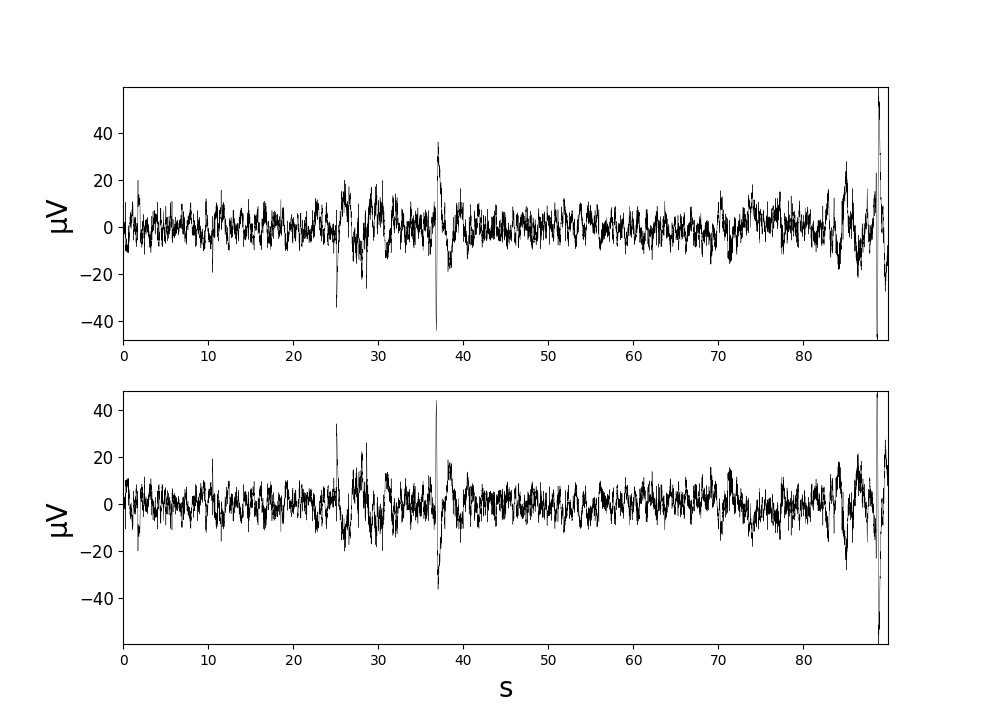}
  \caption{\textit{(top)} 90-s EEG raw signal. \textit{(bottom)} 90-s EEG vertically flipped.}
  \label{fig:flipping}
\end{figure}

The architecture is trained end-to-end via backpropagation, using the \textit{sequence-to-epoch} learning approach. 
Classification algorithms learn to predict the most represented class in the training set, leading to the so called class imbalance problem. Here the least represented classes are balanced by using two techniques:
\textit{(i) data augmentation}, by flipping vertically the data input (i.e. multiply by $-1$ the original signal, see Fig.~\ref{fig:flipping}) belonging to the least represented classes, then \textit{(ii) oversampling} randomly the data so that all the sleep stages are equal in number to the most represented class. In our model, the input is a sequence of three 30-second epochs, and the output is the corresponding target of the central epoch at time $(t)$. So, we refer to the target of the central epoch to compute the most or least represented classes.\\
The model is trained using mini-batch Adam gradient-based optimizer \cite{kingma2014} with a learning rate $lr$. The training procedure runs up to a maximum number of iterations, as long as the break early stopping condition is satisfied - further details in the next subsection \ref{regularizer}.

\subsection{Regularization Techniques and Training Parameters}
\label{regularizer}

\textit{Dropout}. Commonly used as regularizer in convolutional neural networks, it prevents overfitting and co-adaptation of the feature maps \cite{srivastava2014}. During the training procedure a certain number of neurons are randomly removed, dropping units  with a probability $p$. We fix the probability of dropping a connection equal to $50\%$, i.e. $\textit{p} = 0.5$.

\textit{Early stopping}. It provides guidance on how many iterations can be run before the model begins to overfit \cite{prechelt1998}. The training procedure should be stopped as soon as the performance (i.e. F1-score) on the validation set is lower than it was in the previous iteration step. However, in this study, before hastily stopping the learning procedure, the algorithm runs for an additional number of iterations (by fixing the so called \textit{patience} parameter). The model with the highest performance is the one we finally save.

\textit{L2 weight decay}. This technique simply adds a term to the loss function that penalizes the weight values; by doing so it avoids the exploding gradient phenomena \cite{goodfellow2016}. The $lambda$ defines the degree of penalty and it has been set to $10\textsuperscript{-3}$. \\

All the training parameters are fixed as in \cite{supratak2017deepsleepnet}. The Adam optimizer's parameters $beta1$ and $beta2$ have been set to $0.9$ and $0.999$ respectively. The mini-batch size has been set to $100$. During the batch normalization procedure, the $\epsilon$ value of $10\textsuperscript{-5}$ has been added to the mini-batch variance. In order to compute the mean and variance of the training samples, the moving average has been implemented using a fixed decay rate value of $0.999$. The learning rates parameter $lr$ has been fixed to $10\textsuperscript{-4}$. The maximum number of iterations has been set to $100$, with the early stopping \textit{patience} parameter equal to $50$.


\section{Model Calibration}
\label{calibration}

Along with the estimated sleep stage, the model should also provide a calibrated confidence - i.e. the probability associated to the predicted stage should mirror its ground truth correctness likelihood. We adopted label smoothing \cite{szegedy2016} to calibrate our model. It has been shown to be a suitable technique to improve model calibration \cite{muller2019}.\\
In a standard training of a neural network, the cross-entropy loss is minimized using the hard targets $y\textsubscript{k}$ (i.e. hot encoded targets, `$1$' for the correct class and `$0$' for the other). For a network trained with label smoothing, the hard targets are weighted with the \textit{uniform distribution} \textit{$1/K$} (eq. \ref{eq_net6}), and the cross-entropy loss is minimized using the weighted mixture of the targets (eq. \ref{eq_net7}).

\begin{equation}
y\textsubscript{k}\textsuperscript{\textit{LS}\textsubscript{u}} = y\textsubscript{k}\cdot(1-\alpha) + \alpha/K
\label{eq_net6}
\end{equation}

\begin{equation}
\textit{H}(\textbf{\textit{y}}, \textbf{\textit{p}}) = \sum_{k=1}^{K} -y\textsubscript{k}\textsuperscript{\textit{LS}\textsubscript{u}}\cdot\log(\hat{p}\textsubscript{k})
\label{eq_net7}
\end{equation}

where $\alpha$ is the smoothing parameter, $K$ is the total number of classes, $y\textsubscript{k}\textsuperscript{\textit{LS}\textsubscript{u}}$ the targets smoothed with the \textit{uniform distribution}, and $\hat{p}\textsubscript{k}$ the softmax output probabilities.

\subsection{Conditional Probability Distribution in Label Smoothing}
\label{calibration_conditional}

In our study, we introduce a new distribution to smooth the labels, by mainly taking into account the importance in sleep scoring of the transitions from one sleep stage to the other. The idea is to compute the \textit{conditional probability distribution} over the five sleep stages of all the sequences of epochs:

\begin{equation}
\textit{\textbf{M}} = P(stage(t)|stage(t-1), stage(t+1))
\label{eq_net8}
\end{equation}

where in \textit{\textbf{M}} we have the conditional probabilities values for each possible combination of sequences of three sleep stages.
In detail, we compute the probability to be in a stage at time $t$ given the previous $(t-1)$ and the next $(t+1)$ sleep stages over the whole database. The matrix $\textit{\textbf{M}}$ is $K$$\times$$K$$\times$$K$ dimensional, where $K$ is the total number of sleep stages.\\
As stated previously, the architecture takes in input a sequence of three epochs, and outputs the corresponding target of the central epoch $y\textsubscript{k,(t)}$. So, during the training procedure, given the knowledge of the sleep stage at time $(t-1)$ and the sleep stage at time $(t+1)$, the hot encoded $y\textsubscript{k,(t)}$ will be smoothed with the corresponding conditional probability vector from \textbf{M}.

\begin{table}[t]
\caption{Conditional probabilities values computed over the sequences, extracted from the Sleep-EDF v1-2013 dataset, with the label at time $(t-1)$ fixed in awake. \textit{i.e. $\textbf{M}$\textsubscript{$W$,$K$$\times$$K$}}}
\label{conditional_prob}
\begin{center}
\begin{tabular}{|c|c|c|c|c|c|}
\hline
\textit{$W$(t-1)} & \textit{$W$(t+1)} & \textit{$N1$(t+1)} & \textit{$N2$(t+1)} & \textit{$N3$(t+1)} & \textit{$R$(t+1)} \\
\hline
\textit{$W$(t)} & 0.991 & \cellcolor{green!8} 0.503 & 0.131 & 0.333 & 0.217 \\
\hline
\textit{$N1$(t)} & 0.008 & \cellcolor{green!8} 0.495 & 0.581 & 0.000 & 0.109 \\
\hline
\textit{$N2$(t)} & 0.000 & \cellcolor{green!8} 0.002 & 0.275 & 0.000 & 0.000 \\
\hline
\textit{$N3$(t)} & 0.000 & \cellcolor{green!8} 0.000 & 0.006 & 0.667 & 0.000 \\
\hline
\textit{$R$(t)} & 0.000 & \cellcolor{green!8} 0.000 & 0.006 & 0.000 & 0.674\\
\hline
\end{tabular}
\end{center}
\end{table}

In Table \ref{conditional_prob} we report an example of the conditional probability values computed over the sequences extracted from the Sleep-EDF v1-2013 dataset (see section \ref{data}), with the label at time $(t-1)$ fixed in sleep stage awake. We highlight in light-green an example of the conditional probability vector to use when we had awake $\textit{W}$ at time $(t-1)$ and $\textit{N1}$ at time $(t+1)$, which results in the following smoothed target:

\begin{equation}
y\textsubscript{k}\textsuperscript{\textit{LS}\textsubscript{s}} = y\textsubscript{k}\cdot(1-\alpha) + \alpha\cdot\textit{\textbf{M}\textsubscript{$W,K,N1$}} 
\label{eq_net9}
\end{equation}

The cross-entropy loss is minimized using the weighted mixture of the hard targets with these conditional probability distributions.\\

The smoothing parameter $\alpha$ for the \textit{uniform distribution} and the \textit{conditional probability distribution} weighting has been set to $0.1$ and $0.2$ respectively. These two values gave us the highest performance. In both, we explored $\alpha$ values up to $0.5$. 

\section{Estimating Uncertainty}
\label{uncertainty}

In order to estimate the model uncertainty, we exploit the \textit{dropout} regularization technique. As explained above, during the training procedure, at each iteration, \textit{dropout} removes a certain number of units within our network at random. It randomly samples a certain number of sub-networks, so that each time the model's architecture is slightly different. In a standard application, \textit{dropout} is used only during the training phase. At test time, instead, all the trained neurons and connections are used - i.e. all the weights of the whole network. The output could be interpreted as an averaging ensemble of all the sub-networks.
We employ, for the first time in sleep staging, the \textit{Monte Carlo} (\textit{MC}) \textit{dropout} \cite{gal2016}, to quantify the model uncertainty, and  to finally enhance the performance of the scoring architecture. Monte Carlo refers to a specific class of algorithms that rely on random sampling, to provide estimates and distributions of numerical quantities. \textit{MC dropout} simply consists in applying the randomized sampling even at test time. The different sub-networks could be interpreted as \textit{Monte Carlo} samples extracted from the space of all the possible models. As a result, by applying \textit{dropout} $N$ times at inference time (with the probability of dropping a connection $\textit{p} = 0.5$), we would get $N$ different predictions. We compute the \textit{mean} and the \textit{variance} of the $N$ predictions for each sleep stage $k$

\begin{equation}
\mu\textsubscript{i,k} = \frac{\sum_{n=1}^{N}\hat{p}\textsubscript{n,i,k}}{N}
\label{eq_net10}
\end{equation}

\begin{equation}
\sigma^2\textsubscript{i,k} = \frac{\sum_{n=1}^{N}(\hat{p}\textsubscript{n,i,k} - \mu\textsubscript{i,k})^2}{N}
\label{eq_net11}
\end{equation}

where $\hat{p}\textsubscript{n,i,k}$ is the output probability for the sleep stage $k$ of the n-th prediction for the input \textbf{x}\textsubscript{i}. The final prediction $\hat{y\textsubscript{i}}$ of the model will be given by $max$(\pmb{$\mu$}\textsubscript{i}).\\

The uncertain predictions will be then estimated by analysing both their computed \textit{mean} and \textit{variance}. The selection procedure of the uncertain sleep stages is explained in detail in subsection \ref{query}. The selected uncertain predictions could be then presented to the physician for a secondary review.

\begin{table}[h]
\caption{Number and percentage of 30-second epochs per sleep stage \\of the Sleep-EDF datasets with different trimming.}
\label{database}
\begin{center}
\begin{scriptsize}
\begin{tabular}{c|ccccc|c}
Dataset & W & N1 & N2 & N3 & R & Total \\
\hline
\noalign{\vskip 1mm}
\multirow[c]{2}{*}{\specialcell[c]{v1-2013\\$\pm$30mins}}    & 8285  & 2804  & 17799  & 5703  & 7717  & \multirow[c]{2}{*}{42308}\\
& \scriptsize(19.6\%)  & \scriptsize(6.6\%)  & \scriptsize(42.1\%)  & \scriptsize(13.5\%)  & \scriptsize(18.2\%)  & \\
\hline
\noalign{\vskip 1mm}
\multirow[c]{2}{*}{v1-2013} & 5907 & 2687 & 17255 & 5465 & 7647 & \multirow[c]{2}{*}{38961}\\
& \scriptsize(15.2\%)  & \scriptsize(6.9\%)  & \scriptsize(44.3\%)  & \scriptsize(14.0\%)  & \scriptsize(19.6\%)  & \\
\hline
\hline
\noalign{\vskip 1mm}
\multirow[c]{2}{*}{\specialcell[c]{v2-2018\\$\pm$30mins}} & 65951  & 21522  & 69132  & 13039  & 25835  & \multirow[c]{2}{*}{195479}\\
& \scriptsize(33.7\%)  & \scriptsize(11.0\%)  & \scriptsize(35.4\%)  & \scriptsize(6.7\%)  & \scriptsize(13.2\%)  & \\
\hline
\noalign{\vskip 1mm}
\multirow[c]{2}{*}{v2-2018} & 43055 & 19168 & 64408 & 12042 & 25275 & \multirow[c]{2}{*}{163948}\\
& \scriptsize(26.3\%)  & \scriptsize(11.7\%)  & \scriptsize(39.3\%)  & \scriptsize(7.3\%)  & \scriptsize(15.4\%)  & \\
\hline
\end{tabular}
\end{scriptsize}
\end{center}
\end{table}

\section{Data}
\label{data}

\textbf{Sleep-EDF (SC)}. The Sleep-EDF Sleep Cassette is a subset of the open source Sleep-EDF dataset\cite{Sleep-EDF}. The PSG data belong to 78 subjects (37 males and 41 females) aged from 25 to 101 years. Except for the first nights of subjects 36 and 52, and for the second night of subject 13, for all the subjects are available two whole nights, resulting in 153 PSG recordings. 
Each recording includes two scalp EEG channels (Fpz-Cz and Pz-Cz), one EOG (horizontal) channel, one submental chin EMG channel and one oro-nasal respiration channel. The recordings are manually scored by sleep experts on 30-second epochs according to Rechtschaffen and Kales scoring rules \cite{rechtschaffen1968manual}, resulting in the eight classes Wake, N1, N2, N3, N4, REM, MOVEMENT and UNKNOWN. In order to use the AASM standard \cite{iber2007aasm}, we have merged the N3 and N4 stages into a single stage N3, and we have excluded the MOVEMENT and UNKNOWN classes. In many recordings there were long wake periods before the patients went to sleep and after they woke up. We have done experiments with the two common ways these periods are trimmed in literature: 1) only \textit{in-bed} parts are employed \cite{imtiaz2015}, i.e. from \textit{light-off} time to \textit{light-on} time; 2) 30 minutes of data before and after \textit{in-bed} parts are taken into account in the experiments \cite{supratak2017deepsleepnet}. In our study we have considered the EEG Fpz-Cz channel, with a sampling rate of 100 Hz and without any pre-processing.\\ 
In order to facilitate the comparison with many existing deep learning based scoring algorithms, in this work we use the last expanded version published in 2018, and also the previous upload of the Sleep-EDF database published in 2013. In the older upload there were only 39 PSG recordings from 20 subjects.
In Table~\ref{database} we report a summary of the total number and percentage of the epochs per sleep stage.

\section{Results}

\subsection{Design of Experiments}

The validation procedure is in line with the state-of-the-art methods considered in Table \ref{SotAvsOur} in subsection \ref{SotA_comparison}. In fact, we evaluate our model using the \textit{k}-fold cross-validation scheme. We set \textit{k} equal to $20$ for v1-2013 and $10$ for v2-2018 Sleep-EDF datasets. In Table \ref{dbSplit} we summarize the data split for each dataset. In our study we decided to further standardize the experiments by considering in each fold the same subject IDs used in \cite{phan2021}. We believe that in such small datasets, the subjects involved in the training/validation/test set may have an impact on the final results.

\begin{table}[h]
\caption{Summary of the Sleep-EDF dataset and the data split.}
\label{dbSplit}
\begin{center}
\begin{tabular}{c|c|c|c|c}
\multicolumn{1}{c}{\multirow[c]{2}{*}{Dataset}} & \multicolumn{1}{c}{\multirow[c]{2}{*}{Size}} & \multicolumn{1}{c}{Experimental} & \multicolumn{1}{c}{Held-out} & \multicolumn{1}{c}{Held-out} \\
& & Setup & Validation Set & Test Set \\
\hline
\noalign{\vskip 1mm}
v1-2013 & 20 & 20-fold CV & 4 subjects & 1 subject\\
\hline
\noalign{\vskip 1mm}
v2-2018 & 78 & 20-fold CV & 7 subjects & 7 subjects\\
\hline
\end{tabular}
\end{center}
\end{table}

The following experiments are conducted:

\begin{itemize}
	\item \textbf{\textit{base}}. The model is trained not considering model calibration, and without label smoothing. 
	\item \textbf{\textit{base+LS}\textsubscript{u}}. The model is trained taking into account the confidence calibration, using \textit{label smoothing} with \textit{uniform distribution} - i.e. the hard targets are weighted with the \textit{uniform distribution}.
	\item \textbf{\textit{base+LS}\textsubscript{s}}. The model is trained taking into account the confidence calibration, using \textit{label smoothing} with our statistical analysis done on the sequences of sleep stages - i.e. the hard targets are weighted with the \textit{conditional probability distribution}.
\end{itemize}

These three models, differently trained, have been evaluated with and without using the \textit{MC dropout} sampling technique. In Table \ref{ECE} subsection \ref{models_analysis} we present the results obtained for the three models, and the impact of \textit{MC dropout} at inference time.\\

The models have been implemented in TensorFlow 1.14, and trained on a single workstation running Ubuntu 18.04.2 with a Intel Core i7-8700K CPU, an NVIDIA GTX 1080 GPU with 8 GB memory and 32 GB RAM memory.

\begin{table}[!t]
\caption{Overall performance and calibration measure of the models obtained from $20$-fold cross-validation with and without \textit{MC} on Sleep-EDF v1-2013 $\pm$30mins dataset. Best shown in bold.}
\label{ECE}
\begin{center}
\begin{tabular}{c|c|cccc|cc}
\multicolumn{1}{c}{\multirow[c]{2}{*}{}} & \multicolumn{1}{c}{} & \multicolumn{4}{c}{Overall Metrics} & \multicolumn{2}{c}{Calibration}  \\
& Models & Acc. & MF1 & \textit{k} & F1 & ECE & \textit{conf}\\
\hline
\noalign{\vskip 1mm}
\multirow[c]{3}{*}{\rotatebox[origin=c]{90}{\textit{w/o MC}}} & \textit{base} & 82.3 & 76.6 & 0.76 & 82.5 & 0.111 & 93.4 \\
& \textit{base+LS}\textsubscript{u} & 82.8 & 77.2 & 0.77 & 83.0 & \textbf{0.023 }& 80.5 \\
& \textit{base+LS}\textsubscript{s} & 82.7 & 76.4 & 0.76 & 82.7 & 0.071 & 89.7 \\
\hline
\noalign{\vskip 1mm}
\multirow[c]{3}{*}{\rotatebox[origin=c]{90}{\textit{w/ MC}}} & \textit{base} & 83.0 & 77.1 & 0.77 & 83.0 & 0.060 & 89.0 \\
& \textbf{\textit{base+LS}\textsubscript{u}} & \textbf{84.0} & \textbf{78.0} & \textbf{0.78} & \textbf{83.9} & 0.055 & 78.5 \\
& \textit{base+LS}\textsubscript{s} & 83.4 & 77.0 & 0.77 & 83.3 & \textbf{0.031} & 86.5 \\
\hline
\end{tabular}
\end{center}
\end{table}

\subsection{Metrics}

\textit{Performance}. The per-class F1-score, the overall accuracy (\textit{Acc.}), the macro-averaging F1-score (\textit{MF1}) and the Cohen's kappa (\textit{k}) have been computed from the predicted sleep stages from all the folds to evaluate the performance of our model \cite{cohen1960, sokolova2009}. In our experiments the weighted-averaging F1-score has been also reported, taking into account also the label imbalance problem. It computes the average of the metric weighted by the number of true instances for each label. The F1-score computed in this way is not a realistic weighted average of the precision and recall, but it takes into account the high imbalance between the sleep stages.

\textit{Calibration}. We evaluated the calibration of our model using the Expected Calibration Error (ECE) proposed in \cite{naeini2015}. It approximates the difference in expectation between accuracy $acc$ and confidence $conf$, where with confidence it refers to the softmax output probabilities.\\
More in detail, we first divide the predictions into $M$ equally spaced bins (size $1/M$), then for each bin we compute the accuracy $acc(B\textsubscript{m})$ and we define the average predicted probability value $conf(B\textsubscript{m})$:

\begin{equation}
acc(B\textsubscript{m}) = \frac{1}{|B\textsubscript{m}|}\cdot\sum_{i \in B\textsubscript{m}}^{K} \textbf{1}(\hat{y\textsubscript{i}} = y\textsubscript{i})
\label{eq_net12}
\end{equation}

\begin{equation}
conf(B\textsubscript{m}) = \frac{1}{|B\textsubscript{m}|}\cdot\sum_{i \in B\textsubscript{m}}^{K} \hat{p\textsubscript{i}}
\label{eq_net13}
\end{equation}

where $\textit{y\textsubscript{i}}$ and $\hat{y\textsubscript{i}}$ are the true and predicted labels for the sample $i$, $B\textsubscript{m}$ is the group of samples whose predicted probability values falls into the interval $I\textsubscript{m} = (\frac{m-1}{M}, \frac{m}{M}]$, and $\hat{p\textsubscript{i}}$ is the predicted probability value for sample $i$.\\
Then we finally compute the weighted average of the $acc$ and $conf$ difference of the $M$ bins, 

\begin{equation}
ECE = \sum_{m=1}^{M} \frac{|B\textsubscript{m}|}{n} \cdot |acc(B\textsubscript{m}) - conf(B\textsubscript{m})|
\label{eq_net14}
\end{equation}

where $n$ is the number of samples in each bin. \\
Clearly, perfectly calibrated models have $acc(B\textsubscript{m}) = conf(B\textsubscript{m})$ for all $m \in \{1,..,M\}$, resulting in $ECE = 0$.

\subsection{Analysis of Experiments}

\label{models_analysis}

\begin{figure}[t]
  \centering
    \adjincludegraphics[width=\linewidth, trim={0 0 {.05\width} 0},clip]{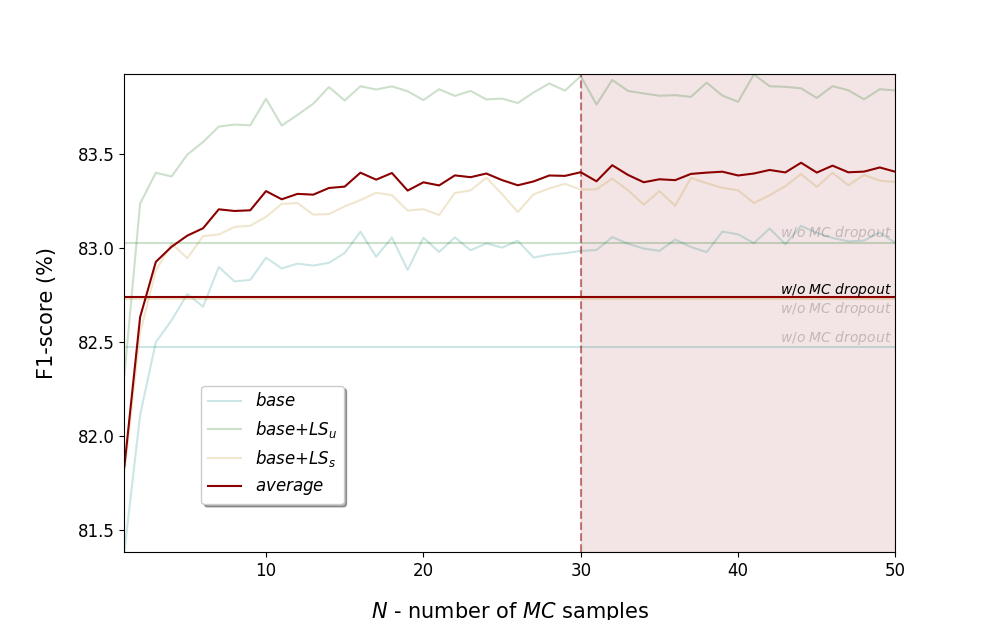}
  \caption{F1-score against the number of \textit{Monte Carlo} samples $N$ of the three models (\textit{base}, \textit{base+LS}\textsubscript{u} and \textit{base+LS}\textsubscript{s}) evaluated on Sleep-EDF v1-2013 dataset. \textit{Monte Carlo} sampling converges after 30 samples without further significant improvement on the average of the three models.}
  \label{fig:MC-samples}
\end{figure}

In table \ref{ECE} we report the overall performance and the calibration measure of three different models, with and without \textit{Monte Carlo dropout} at inference time, to which we refer \textit{w/o MC} and \textit{w/ MC} respectively. In the following, we analyse only the results obtained on the Sleep-EDF v1-2013 $\pm$30mins dataset, since the findings are still valid for its expanded v2-2018 $\pm$30mins version.\\ 
In our tests \textit{w/o MC}, we show the efficiency of label smoothing in calibrating the model. The $conf$ value refers to the average of all the predicted probability values. In both \textit{LS}\textsubscript{u} and \textit{LS}\textsubscript{s} models, the $conf$ probability better reflects the ground truth correctness likelihood - i.e. accuracy value. Indeed, it results in a better ECE value $0.023$ and $0.071$, compared to the higher $0.111$ for the \textit{base} model. The overall performance are preserved or even improved. \\
By using \textit{MC} at test time, we show the efficiency of label smoothing and \textit{MC} techniques in both calibrating and enhancing the performance of the model. It is quite interesting the impact of \textit{MC dropout}: an increase in overall metrics and a decrease in the average predicted  probability values. This justifies a better calibrated model by using our \textit{conditional probability distribution} smoothing technique \textit{LS}\textsubscript{s} - i.e. ECE value equal to $0.031$.

In Fig.~\ref{fig:MC-samples} we report the F1-score against the number of \textit{Monte Carlo} samples $N$, evaluated over all our experiments. 
Interesting how \textit{Monte Carlo} sampling outperforms the experiments done without applying \textit{MC} after approximately three samples, on the average of the three models. On average we get a plateau after 30 samples, so we decided to set $N$ equal to $30$.\\
From here on, all the results will refer to the best of our models \textit{base+LS\textsubscript{u}}, by using \textit{MC} sampling at test time.\\

In Tables \ref{confusion_matrix_v1} and \ref{confusion_matrix_v2} we report the confusion matrix and the per-class performance of the best of our models evaluated on Sleep-EDF v1-2013 $\pm$30mins and v2-2018 $\pm$30mins respectively. The \textit{i-th} row and the \textit{j-th} column indicates the percentage number of 90-s EEG instances with the true label being \textit{i-th} class and the predicted label being \textit{j-th} class. In bold we highlight the percentage number of instances well classified. As expected \cite{rosenberg2013}, the lowest performance has been obtained for the N1 sleep stage, i.e. F1-score 44.4\% and 46.0\%; most of the N1 have been wrongly classified in awake, N2 and REM. The F1-score for all the other sleep stages were in range between 82.4\% and 88.2\% on v1-2013 $\pm$30mins, and between 76.4\% and 91.5\% on v2-2018 $\pm$30mins.

\begin{table}[t]
\caption{Confusion matrix obtained from 20-fold cross-validation \\on Sleep-EDF v1-2013 $\pm$30mins dataset.}
\label{confusion_matrix_v1}
\begin{center}
\begin{tabular}{c|ccccc|ccc}
\multicolumn{1}{c}{\multirow[c]{2}{*}{}} & \multicolumn{5}{c}{Predicted} & \multicolumn{3}{c}{Per-class Metrics} \\
& W & N1 & N2 & N3 & R & Pr. & Rec. & F1 \\
\hline
\noalign{\vskip 1mm}
W & \textbf{90.4} & 6.0 & 1.4 & 0.3 & 1.9 & 84.0 & 90.4 & 87.1 \\
N1 & 18.6 & \textbf{42.6} & 19.2 & 0.8 & 18.8 & 46.5 & 42.6 & 44.4 \\
N2 & 3.0 & 2.4 & \textbf{86.0} & 4.5 & 4.1 & 89.9 & 86.0 & 87.9 \\
N3 & 1.3 & 0.2 & 7.9 & \textbf{90.6} & 0 & 85.9 & 90.6 & 88.2 \\
R & 3.5 & 5.7 & 7.8 & 0.1 & \textbf{82.9} & 82.0 & 82.8 & 82.4 \\
\hline
\end{tabular}
\end{center}
\end{table}

\begin{table}[t]
\caption{Confusion matrix obtained from 10-fold cross-validation \\on Sleep-EDF v2-2018 $\pm$30mins dataset.}
\label{confusion_matrix_v2}
\begin{center}
\begin{tabular}{c|ccccc|ccc}
\multicolumn{1}{c}{\multirow[c]{2}{*}{}} & \multicolumn{5}{c}{Predicted} & \multicolumn{3}{c}{Per-class Metrics} \\
& W & N1 & N2 & N3 & R & Pr. & Rec. & F1 \\
\hline
\noalign{\vskip 1mm}
W & \textbf{90.0} & 7.5 & 0.6 & 0.2 & 1.7 & 93.0 & 90.0 & 91.5 \\
N1 & 14.2 & \textbf{48.1} & 24.5 & 1.0 & 12.2 & 44.1 & 48.1 & 46.0 \\
N2 & 0.7 & 8.5 &\textbf{ 80.3} & 5.3 & 5.2 & 85.6 & 80.3 & 82.9 \\
N3 & 0.2 & 0.3 & 12.8 & \textbf{86.5} & 0.2 & 73.0 & 86.5 & 79.2 \\
R & 3.4 & 8.8 & 7.6 & 0.8 & \textbf{79.4} & 73.7 & 79.4 & 76.4 \\
\hline
\end{tabular}
\end{center}
\end{table}

\subsection{Uncertainty Estimate}
\label{query}

\begin{figure*}
\centering
\adjincludegraphics[width=0.9\textwidth, trim={{.08\width} 0 {.08\width} 0},clip]{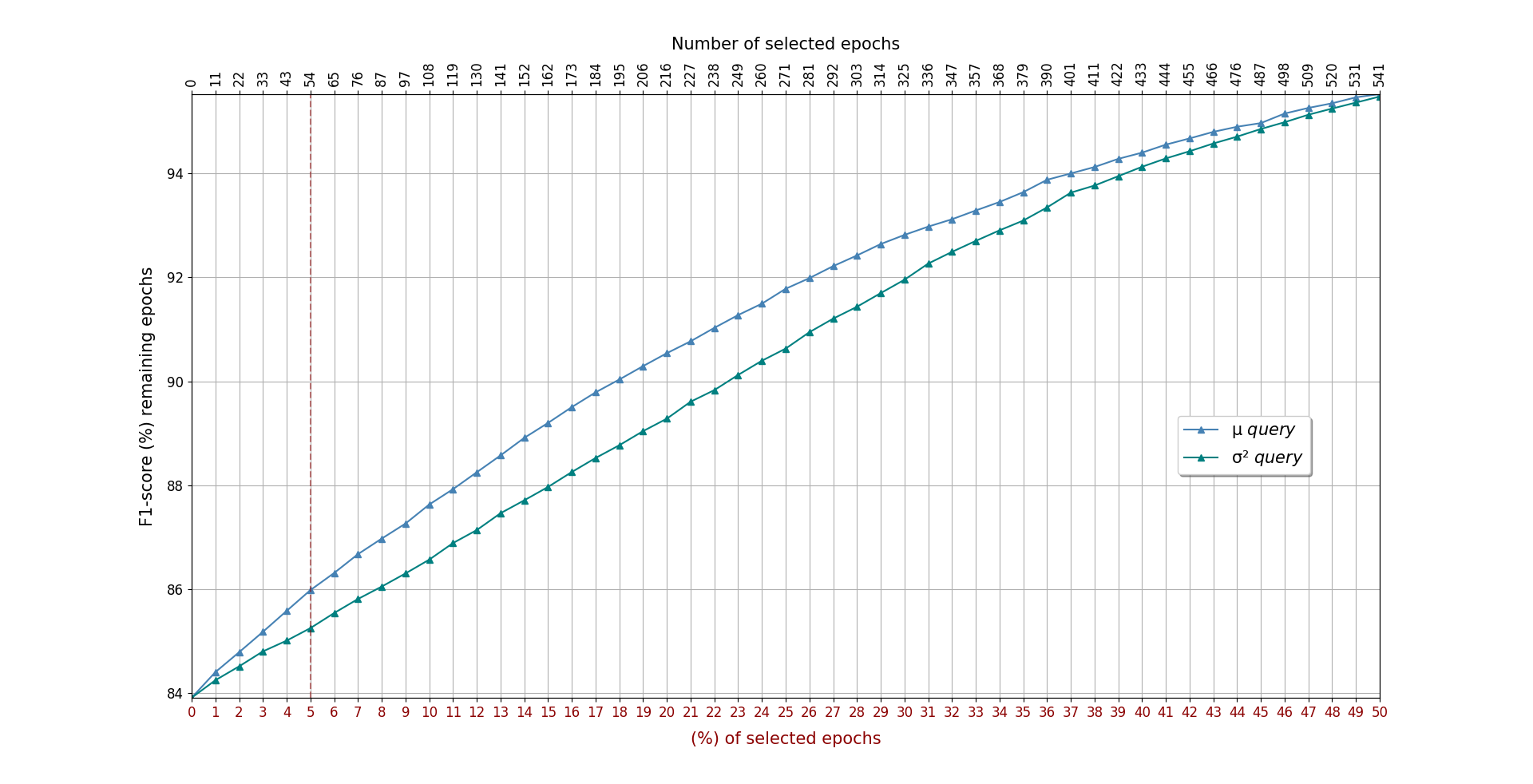}
\caption{F1-score computed over the remaining epochs after the \textit{query} procedure against the percentage number of epochs to select. In light green and in light blue the F1-score performance in case the selection procedure has been done using the variance ($\sigma^2$ \textit{query}) and the mean ($\mu$ \textit{query}) respectively. The performance refers to the best of our model evaluated on Sleep-EDF v1-2013 $\pm$30mins dataset.}
\label{fig:F1score_Query}
\end{figure*}

\begin{figure*}
\centering
\adjincludegraphics[width=0.9\textwidth, trim={{.08\width} 0 {.08\width} 0},clip]{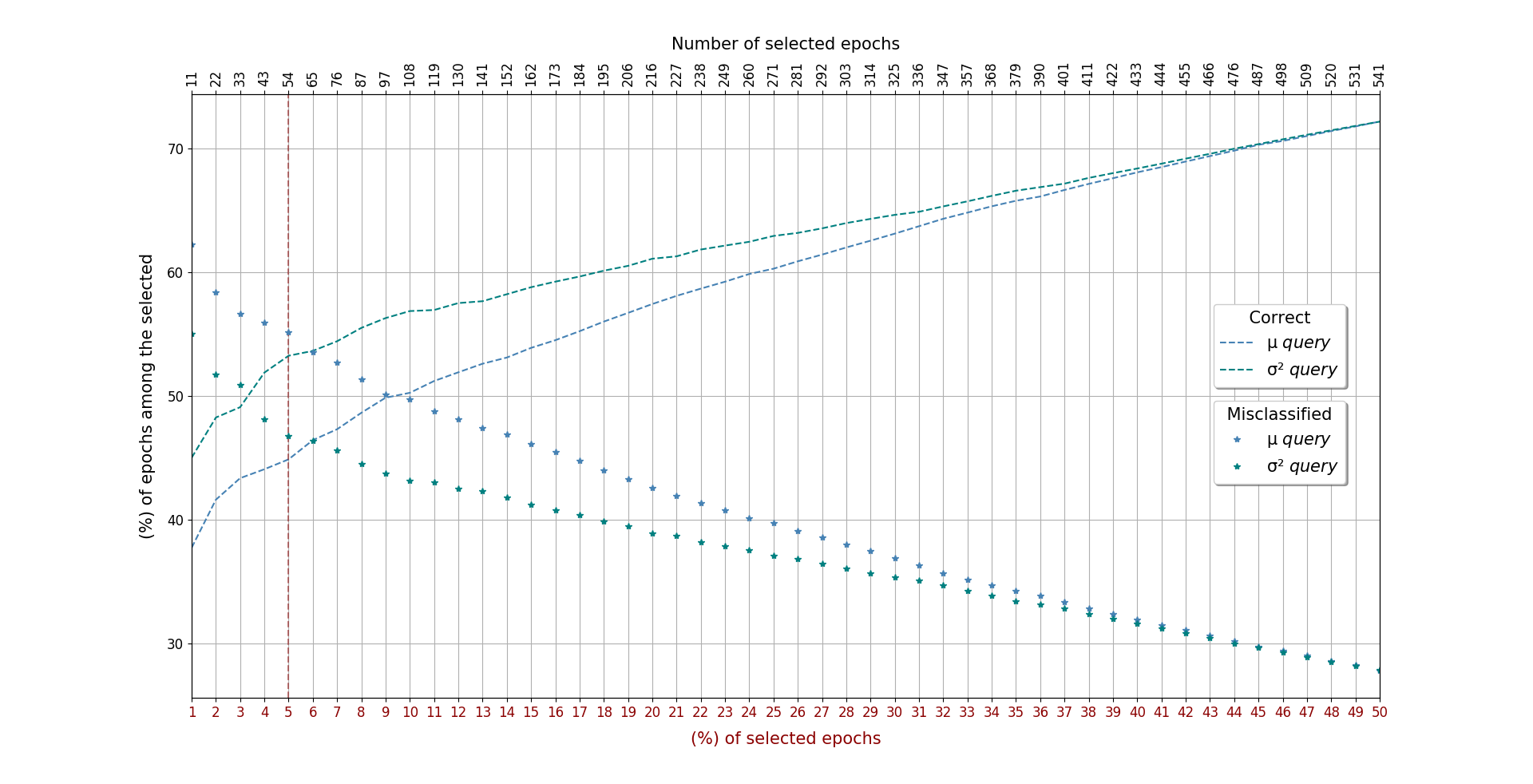}
\caption{Percentage of misclassified and correctly classified epochs among the \textit{q}\% selected. In light green and in light blue the  percentage values in case the selection procedure has been done using the variance ($\sigma^2$ \textit{query}) and the mean ($\mu$ \textit{query}) respectively. The performance refers to the best of our models evaluated on Sleep-EDF v1-2013 $\pm$30mins dataset.}
\label{fig:Corr-Miscl_Query}
\end{figure*}

\textit{MC dropout} enables the estimate of the uncertain predictions. In order to select the uncertain instances, at first, we used the \textit{variance} ($\sigma^2$ of the predicted probability values obtained from the $N$ sampling). The selection procedure (also referred to as \textit{query} procedure) simply rely on the setting of a threshold value \textit{q}\%, that corresponds to the percentage number of epochs - for each PSG recording - to select and to send potentially to the physician for a secondary review. The epochs with the highest values of \textit{variance} will be the \textit{q}\% selected. We also tried to use the \textit{mean} ($\mu$ of the predicted probability values obtained from the $N$ sampling) to select the uncertain instances. In this case the epochs with the lowest \textit{mean} values will be the \textit{q}\% selected.\\
The selected epochs, in both cases, correspond to the predictions where the averaging ensemble of the models outputs the higher uncertainty. 
In Fig.~\ref{fig:F1score_Query} we report the F1-score computed over the remaining epochs against the percentage number of selected instances. We have fixed the \textit{q}\% threshold value to $5\%$, because it was considered to be a reasonable number of epochs ($54$ on average for each PSG recording) to select and to eventually present to the physician for a secondary review. The results show that by using $\mu$ in the selection procedure we obtain higher performance. In Fig.~\ref{fig:Corr-Miscl_Query} we also report, for each \textit{q}\% number of selected instances, the percentage of misclassified and correctly classified epochs among the selected ones. As illustrated, by using $\mu$, the percentage number of misclassified epochs are greater than the correctly classified up to the selection threshold \textit{q}\% equal to $10\%$. Whilst, by using $\sigma^2$, the percentage number of selected epochs \textit{q}\% radically decreases to $2\%$.\\
In Table \ref{uncertainty_var_mean} we also report the average of the per-class $\sigma^2$ and $\mu$ predicted probability values, to have an overall estimate of the model uncertainty, evaluated on both Sleep-EDF v1-2013 $\pm$30mins and v2-2018 $\pm$30mins datasets. As expected, the results show that the model has more difficulty in classifying N1 and REM epochs, while providing greater confidence in classifying W, N2 and N3 sleep stages (lower variance and higher predicted probability values).

\begin{table}[ht]
\caption{Per-class $\sigma^2$ and $\mu$ of the predicted probability \\ values computed on Sleep-EDF v1-2013 $\pm$30mins \\ and v2-2018 $\pm$30mins datasets.}
\label{uncertainty_var_mean}
\begin{center}
\begin{scriptsize}
\begin{tabular}{c|c|ccccc}
Dataset & Total & W & N1 & N2 & N3 & R\\
\hline
\noalign{\vskip 1mm}
\multirow[c]{2}{*}{{\specialcell[c]{v1-2013\\$\pm$30mins}}} &$\sigma^2$ & 0.008 & 0.024 & 0.009 & 0.006 & 0.014 \\
&$\mu$& 81.3 & 60.2 & 80.4 & 81.7 & 75.3 \\
\hline
\noalign{\vskip 1mm}
\multirow[c]{2}{*}{{\specialcell[c]{v2-2018\\$\pm$30mins}}} &$\sigma^2$ & 0.005 & 0.015 & 0.009 & 0.006 & 0.013 \\
&$\mu$& 82.9 & 60.4 & 74.7 & 79.5 & 70.4 \\
\hline
\end{tabular}
\end{scriptsize}
\end{center}
\end{table}

\subsection{Comparison with state-of-the-art}
\label{SotA_comparison}

\begin{table*}[!t]
\caption{Comparison between our method and the other deep learning-based automatic sleep scoring systems using raw single channel Fpz-Cz, evaluated on Sleep-EDF datasets with overall accuracy (Acc.), macro F1-score (MF1), Cohen’s Kappa (k) and per-class F1-score. The best performance metrics for each dataset are indicated in bold.}
\label{SotAvsOur}
\begin{center}
\resizebox{15cm}{!}{\begin{tabular}{c|c|c|c|ccc|ccccc}
\multicolumn{1}{c}{\multirow[c]{2}{*}{Datasets}} & \multicolumn{1}{c}{\multirow[c]{2}{*}{Methods}} & \multicolumn{1}{c}{Training} & \multicolumn{1}{c}{Sequences} & \multicolumn{3}{c}{Overall Metrics} & \multicolumn{5}{c}{Per-Class F1-Score}\\
& & Param. & of epochs & Acc. & MF1 & \textit{k} & W & N1 & N2 & N3 & R\\
\hline
\noalign{\vskip 1mm}
\multirow{9}{*}{{\specialcell[c]{v1-2013 \\ $\pm$30mins}}} & FCNN+RNN \cite{phan2021} & $\sim$ 5.6M & 20 & 81.8 & 75.6 & 0.75 & 89.4 & 44.1 & 84.0 & 84.0 & 76.3 \\
& DeepSleepNet \cite{supratak2017deepsleepnet} & $\sim$ 24.7M & 25 & 81.4 & 75.6 & 0.75 & 83.9 & 43.5 & 85.4 & 84.1 & 81.1\\
& DeepSleepNet \cite{supratak2017deepsleepnet} $\dagger$ & $\sim$ 24.7M & 25 & 82.0 & 76.9 & 0.76 & 84.7 & 46.6 & 85.9 & 84.8 & 82.4\\
& IITNet \cite{back2019} $\dagger$ & - & 10 & 84.0 & 77.7 & 0.78 & 87.9 & 44.7 & 88.0 & 85.7 & 82.1\\
& Our method & $\sim$ 0.6M & \textbf{3} & 84.0 & 78.0 & 0.78 & 87.1 & 44.4 & 87.9 & 88.2 & 82.4\\
& SleepEEGNet \cite{mousavi2019}  $\dagger$ & $\sim$ 2.6M & 10 & 84.3 & 79.7 & 0.79 & 89.2 & \textbf{52.2} & 86.8 & 85.1 & \textbf{85.0}\\
& SeqSleepNet+ \cite{phan2020} & \textbf{$\sim$ 0.2M} & 20 & 85.2 & 78.4 & 0.80 & 90.5 & 45.4 & \textbf{88.1} & 86.4 & 81.8\\
& Naive Fusion \cite{phan2021} & $\sim$ 5.8M & 20 & 85.0 & 78.8 & 0.79 & 91.7 & 48.8 & 87.2 & 82.9 & 83.6\\
& TinySleepNet \cite{supratak2020}  $\dagger$ & $\sim$ 1.3M & 15 & 85.4 & 80.5 & 0.80 & 90.1 & 51.4 & 88.5 & \textbf{88.3} & 84.3\\
& XSleepNet2 \cite{phan2021} & $\sim$ 5.8M & 20 & \textbf{86.3} & \textbf{80.6} & \textbf{0.81} & \textbf{92.2} & 51.8 & 88.0 & 86.8 & 83.9\\
 \hline
\noalign{\vskip 1mm}
\multirow{6}{*}{{\specialcell[c]{v1-2013}}} & Naive Fusion \cite{phan2021} & $\sim$ 5.8M & 20 & 80.2 & 74.9 & 0.72 & 77.3 & 47.4 & 85.8 & 84.8 & 79.3\\
& DeepSleepNet \cite{supratak2017deepsleepnet} & $\sim$ 24.7M & 25 & 82.5 & 76.8 & 0.76 & 80.1 & 47.3 & 87.0 & 85.7 & 83.8\\
& DeepSleepNet \cite{supratak2017deepsleepnet} $\dagger$& $\sim$ 24.7M & 25 & 82.6 & 77.1 & 0.76 & \textbf{82.9} & 46.8 & 86.5 & 84.1 & 85.2\\
& FCNN+RNN \cite{phan2021} &  $\sim$ 5.6M & 20 & 81.3 & 76.0 & 0.74 & 76.4 & 50.0 & 86.8 & 85.3 & 81.3\\
& SeqSleepNet+ \cite{phan2020} & \textbf{$\sim$ 0.2M} & 20 & 82.2 & 74.1 & 0.75 & 78.5 & 37.1 & 87.6 & 86.2 & 81.2\\
& Our method & $\sim$ 0.6M & \textbf{3} & 82.6 & 76.3 & 0.76 & 81.6 & 42.4 & 87.4 & \textbf{87.9} & 82.1\\
& XSleepNet2 \cite{phan2021} & $\sim$ 5.8M & 20 & \textbf{83.9} & \textbf{78.7} & \textbf{0.77} & 81.6 & \textbf{52.9} & \textbf{88.1} & 85.3 & \textbf{85.4}\\
\hline
\hline
\noalign{\vskip 1mm}
\multirow{8}{*}{{\specialcell[c]{v2-2018 \\ $\pm$30mins}}} & DeepSleepNet \cite{supratak2017deepsleepnet} & $\sim$ 24.7M & 25 & 76.9 & 70.7 & 0.69 & 90.8 & 44.8 & 78.5 & 67.9 & 71.3\\
& DeepSleepNet \cite{supratak2017deepsleepnet} $\dagger$ & $\sim$ 24.7M & 25 & 77.1 & 71.2 & 0.69 & 90.4 & 46.0 & 79.1 & 68.6 & 71.8\\
& SleepEEGNet \cite{mousavi2019} $\dagger$ & $\sim$ 2.6M & 10 & 80.0 & 73.6 & 0.73 & 91.7 & 44.1 & 82.5 & 73.5 & 76.1\\
& Our method & $\sim$ 0.6M & \textbf{3} & 80.3 & 75.2 & 0.73 & 91.5 & 46.0 & 82.9 & 79.2 & 76.4\\
& Naive Fusion \cite{phan2021} & $\sim$ 5.8M & 20 & 82.3 & 76.2 & 0.75 & 93.2 & 49.6 & \textbf{86.2} & 79.4 & \textbf{82.5}\\
& SeqSleepNet+ \cite{phan2020} &  \textbf{$\sim$ 0.2M} & 20 & 82.6 & 76.4 & 0.76 & 92.2 & 47.8 & 84.9 & 77.2 & 79.9\\
& FCNN+RNN \cite{phan2021} & $\sim$ 5.6M & 20 & 82.8 & 76.6 & 0.76 & 92.5 & 47.3 & 85.0 & 79.2 & 78.9\\
& TinySleepNet \cite{supratak2020}  $\dagger$ & $\sim$ 1.3M & 15 & 83.1 & \textbf{78.1} & 0.77 & 92.8 & \textbf{51.0} & 85.3 & \textbf{81.1} & 80.3\\
& XSleepNet2 \cite{phan2021} & $\sim$ 5.8M & 20 & \textbf{84.0} & 77.9 & \textbf{0.78} & \textbf{93.3} & 49.9 & 86.0 & 78.7 & 81.8\\
\hline
\noalign{\vskip 1mm}
\multirow{6}{*}{{\specialcell[c]{v2-2018}}} & DeepSleepNet \cite{supratak2017deepsleepnet} & $\sim$ 24.7M & 25 & 76.0 & 72.2 & 0.68 & 88.1 & 45.8 & 79.7 & 74.3 & 72.9\\
& DeepSleepNet \cite{supratak2017deepsleepnet} $\dagger$ & $\sim$ 24.7M & 25 & 76.6 & 73.0 & 0.69 & 88.3 & 46.1 & 79.9 & 76.2 & 74.4\\
& SeqSleepNet+ \cite{phan2020} & \textbf{$\sim$ 0.2M} & 20 & 79.0 & 74.6 & 0.71 & 83.2 & 46.8 & 85.5 & 76.3 & 81.0\\
& Our method & $\sim$ 0.6M & \textbf{3} & 79.0 & 75.1 & 0.72 & \textbf{89.3} & 46.9 & 83.3 & 78.9 & 77.1\\
& Naive Fusion \cite{phan2021} & $\sim$ 5.8M & 20 & 79.1 & 75.1 & 0.71 & 83.9 & 47.8 & 85.4 & 78.4 & 79.8\\
& FCNN+RNN \cite{phan2021} & $\sim$ 5.6M & 20 & 79.3 & 75.1 & 0.71 & 84.2 & 49.1 & 85.2 & 76.8 & 80.4\\
& XSleepNet2 \cite{phan2021} & $\sim$ 5.8M & 20 & \textbf{80.3} & \textbf{76.4} & \textbf{0.73} & 85.2 & \textbf{49.4} & \textbf{86.0} & \textbf{79.8} & \textbf{81.7}\\
\hline
\end{tabular}}
\end{center}
\end{table*}

\begin{table*}[!t]
\caption{Comparison among our methods using raw single channel Fpz-Cz, evaluated on Sleep-EDF datasets with overall accuracy (Acc.), macro F1-score (MF1), Cohen’s Kappa (k), weighted-averaging F1-score (F1) and per-class F1-score. The best performance metrics for each dataset are indicated in bold.}
\label{Query}
\begin{center}
\begin{tabular}{c|c|c|cccc|ccccc}
\multicolumn{1}{c}{\multirow[c]{2}{*}{Datasets}} & \multicolumn{1}{c}{Selection} & \multicolumn{1}{c}{Evaluated} & \multicolumn{4}{c}{Overall Metrics} & \multicolumn{5}{c}{Per-Class F1-Score}\\
& Procedure & Epochs & Acc. & MF1 & \textit{k} & F1 & W & N1 & N2 & N3 & R\\
\hline
\noalign{\vskip 1mm}
\multirow{5}{*}{{\specialcell[c]{v1-2013 \\ $\pm$30mins}}} & - & all & 84.0 & 78.0 & 0.78 & 83.9 & 87.1 & 44.4 & 87.9 & 88.2 & 82.4\\
\cline{2-12}
\noalign{\vskip 0.2mm}
& \multirow[c]{2}{*}{$\sigma^2$ \textit{query}} & kept & 85.7 & 77.9 & 0.80 & 85.2 & 88.6 & 39.6 & 88.9 & 88.5 & 84.2\\
& & rejected & 53.0 & 47.5 & 0.36 & 52.4 & 38.7 & 54.9 & 53.3 & 32.3 & 58.1\\
\cline{2-12}
\noalign{\vskip 0.2mm}
& \multirow[c]{2}{*}{$\mu$ \textit{query}} & kept & \textbf{86.1} & \textbf{79.6} & \textbf{0.81} & \textbf{86.0} & \textbf{89.1} & \textbf{45.7} & \textbf{89.4} & \textbf{89.0} & \textbf{84.8}\\
& & rejected & 44.7 & 43.4 & 0.28 & 44.8 & 42.8 & 39.3 & 49.2 & 37.9 & 47.8\\
\hline
\noalign{\vskip 1mm}
\multirow{5}{*}{{\specialcell[c]{v1-2013}}} & - & all & 82.6 & 76.3 & 0.76 & 82.4 & 81.6 & 42.4 & 87.4 & 87.9 & 82.1\\
\cline{2-12}
\noalign{\vskip 0.2mm}
& \multirow[c]{2}{*}{$\sigma^2$ \textit{query}} & kept & 84.3 & 76.4 & 0.78 & 83.8 & 83.5 & 37.9 & 88.4 & 88.3 & 83.6\\
& & rejected & 50.0 & 45.2 & 0.32 & 49.4 & 39.6 & 51.9 & 47.2 & 30.6 & 56.8\\
\cline{2-12}
\noalign{\vskip 0.2mm}
& \multirow[c]{2}{*}{$\mu$ \textit{query}} & kept & \textbf{84.5} & \textbf{77.8} & \textbf{0.78} & \textbf{84.4} & \textbf{83.6} & \textbf{43.5} & \textbf{88.8} & \textbf{88.7} & \textbf{84.4}\\
& & rejected & 45.6 & 45.7 & 0.29 & 45.8 & 45.9 & 37.6 & 52.2 & 49.3 & 43.5\\
\hline
\hline
\noalign{\vskip 1mm}
\multirow{5}{*}{{\specialcell[c]{v2-2018 \\ $\pm$30mins}}} & - & all & 80.3 & 75.2 & 0.73 & 80.6 & 91.5 & 46.0 & 82.9 & 79.2 & 76.4\\
\cline{2-12}
\noalign{\vskip 0.2mm}
& \multirow[c]{2}{*}{$\sigma^2$ \textit{query}} & kept & 81.7 & 75.9 & 0.75 & 81.8 & 92.3 & 45.6 & 84.0 & 79.9 & 77.5\\
& & rejected & 55.2 & 51.0 & 0.40 & 54.4 & 49.9 & 49.0 & 48.1 & 39.5 & 68.6\\
\cline{2-12}
\noalign{\vskip 0.2mm}
& \multirow[c]{2}{*}{$\mu$ \textit{query}} & kept & \textbf{82.3} & \textbf{76.7} & \textbf{0.76} & \textbf{82.5} & \textbf{92.6} & \textbf{47.1} & \textbf{84.4} & \textbf{80.1} & \textbf{79.4}\\
& & rejected & 42.8 & 41.8 & 0.24 & 43.0 & 44.3 & 39.4 & 45.8 & 35.6 & 43.8\\
\hline
\noalign{\vskip 1mm}
\multirow{5}{*}{{\specialcell[c]{v2-2018}}} & - & all & 79.0 & 75.1 & 0.72 & 79.3 & 89.3 & 46.9 & 83.3 & 78.9 & 77.1\\
\cline{2-12}
\noalign{\vskip 0.2mm}
& \multirow[c]{2}{*}{$\sigma^2$ \textit{query}} & kept & 80.2 & 75.8 & 0.73 & 80.4 & 90.0 & 46.9 & 84.3 & 79.8 & 77.8\\
& & rejected & 57.1 & 52.1 & 0.42 & 56.4 & 48.9 & 47.7 & 50.7 & 41.4 & 71.7\\
\cline{2-12}
\noalign{\vskip 0.2mm}
& \multirow[c]{2}{*}{$\mu$ \textit{query}} & kept & \textbf{80.9} & \textbf{76.7} & \textbf{0.74} & \textbf{81.2} & \textbf{90.7} & \textbf{48.0} & \textbf{84.7} & \textbf{79.8} & \textbf{79.7}\\
& & rejected & 42.4 & 41.1 & 0.23 & 42.6 & 41.3 & 39.7 & 44.9 & 34.6 & 44.9\\
\hline
\end{tabular}
\end{center}
\end{table*}

In Table \ref{SotAvsOur} we compare our best model with the other state-of-the-art methods evaluated on the two versions of the Sleep-EDF database. We report the results for each experimental scenario: 1) only \textit{in-bed} recordings; 2) additional 30 minutes recordings before and after \textit{in-bed}. We have considered only the methods using deep learning based architectures, raw single channel Fpz-Cz, same evaluation procedure (i.e. k-fold cross-validation) and using independent training and test sets. We decided to further standardize our experiments by considering in each fold the same subject IDs used in \cite{phan2021}. All the results indicated by $\dagger$ are not directly comparable, since they use a different set of subject IDs in their training/ evaluation/ testing procedure. The sleep scoring algorithms are compared across the overall metrics (Acc., MF1, Cohen’s Kappa and F1-score) and the per-class F1-score. The proposed DeepSleepNet-Lite achieves slightly lower performance, if not on par, compared to the state-of-the art models on all the Sleep-EDF datasets. The results confirm what we had already partially observed in \cite{fiorillo2020EMBC} on the Sleep-EDF v1-2013: the first \textit{epoch processing block} from DeepSleepNet, trained with a small temporal context in input, still succeed in solving the classification task on the small-sized database. Indeed, on both v1-2013 and v2-2018 \textit{in-bed} recordings, our model achieves an overall accuracy only below 1.3\% compared to the recent state-of-the-art XSleepNet2 \cite{phan2021}. We are not surprised to see our lighter architecture to overperform DeepSleepNet: one of the reasons could be that in \cite{supratak2017deepsleepnet} they have not implemented any early stopping procedure, and they save their model only at the latest iteration step, thus not mitigating the overfitting phenomenon. The number of training parameters of our lighter model are significantly reduced, $\sim$0.6M compared to the others TinySleepNet \cite{supratak2020} $\sim$1.3M, SleepEEGNet \cite{mousavi2019} $\sim$2.6M, FCNN+RNN \cite{phan2021} $\sim$5.6M, Naive Fusion and XSleepNet2 $\sim$5.8M \cite{phan2021} and DeepSleepNet \cite{supratak2017deepsleepnet} $\sim$24.7M. Nevertheless, SeqSleepNet \cite{phan2020} is still the network with the lowest number of parameters $\sim$0.2M. We did not report the number of training parameters for IITNet \cite{back2019} since it was not available in literature.

\subsection{Comparison among our methods}

In Table \ref{Query} we report the results of our best model evaluated on the two versions of the Sleep-EDF database - in both experimental scenarios. The outcomes refer to the performance of the model evaluated before the selection procedure and after the selection procedure, by using $\sigma^2$ and $\mu$ \textit{query} values. We report the results obtained after the selection procedure on both the kept and rejected set of epochs. As a consequence of what we have observed in Fig.~\ref{fig:Corr-Miscl_Query}, on both Sleep-EDF v1-2013 and v2-2018, the model shows an increase in performance over the kept epochs, and a significant decrease on the rejected epochs (below 50\% by using $\mu$ \textit{query}). These results highlight the efficiency of the \textit{query} procedure to select a larger number of misclassified epochs among the selected one. The best performance for each dataset are indicated in bold. We obtain an overall accuracy equal to 86.1\% on v1-2013 $\pm$30mins (84.5\% on \textit{in-bed} only) and equal to 82.3\% on v2-2018 $\pm$30mins 80.9\% on \textit{in-bed} only).

\section{Discussion}

Our simplified deep learning approach to sleep scoring achieves performance slightly lower, if not on par, compared to the existing state-of-the-art methodologies evaluated on the Sleep-EDF database. Beside being trained on a small number of parameters, our method does not require any extra resources to buffer the sequences in input, since it processes sequences of only 90-seconds EEG. Therefore, we may assume that an automatic sleep scoring system does not necessarily have to encode such long temporal structures, rather intrinsic patterns of short-term PSG recordings may be sufficient. \\ However, as a result of further experiments carried out on larger and more heterogeneous databases (e.g. Physio2018 \cite{goldberger2000, ghassemi2018} and SHHS \cite{zhang2018, quan1997}), we can state that these observations are valid on small-sized dataset (i.e. low heterogeneity between subjects).

The major advantage of the proposed approach is that it also provides an estimate of the model uncertainty by exploiting existing layers of the architecture. Unlike the existing confidence estimation algorithms for sleep scoring \cite{patanaik2018end,stephansen2018}, the \textit{Monte Carlo dropout} is easy to implement and it does not require any additional computation over the baseline architecture. Moreover, it produces interpretable outputs, i.e. \textit{mean} and \textit{variance} of the predicted probability values. A clear disadvantage for this approach - as for other ensemble learning based algorithms - is that it needs to be executed N times, obviously increasing the evaluation time by N. However, in a real-time application, it may still be a valid solution because the evaluation of a single sequence takes only a few milliseconds.

The results obtained in subsection \ref{models_analysis}, in case our model is trained by smoothing the labels through the \textit{conditional probability distribution}, are still to be further investigated. The impact of this prior knowledge, inserted during the training of our architecture, is not so obvious. It seems to improve the calibration process of the model while maintaining its overall good performance. Even if with this technique we succeed to better calibrate our network, we do not equally succeed in obtaining higher performance using it in combination with \textit{Monte Carlo dropout}. Therefore, unlike what we expected, it is not always the case that a better calibrated architecture leads to higher performance, or even, to a better estimate of the model uncertainty.

\section{Conclusion and Future Works}

We propose DeepSleepNet-Lite a simplified and lightweight automatic sleep scoring architecture, providing the predicted sleep stages along with an estimate of their uncertainty. The scoring system is based on raw single channel EEG, and it processes 90-seconds time sequences. Although the proposed simple feed forward architecture has proven to be as efficient as RNNs based architectures, we cannot conclude that by using only this first representation learning block we will reach equally good results on larger databases. The \textit{Monte Carlo dropout} technique allows us to enhance the performance of the architecture and to identify a relevant number of misclassified epochs among the ones selected during the query procedure. DeepSleepNet-Lite has a low capacity, i.e. low number of training parameters, hence less prone to overfitting on a small dataset. Therefore the need to further investigate its robustness on larger database. It would be interesting to simulate the query procedure on the recent state-of-the-art architectures, e.g. XSleepNet2, to assess its benefit on them. Our lightweight sleep scoring approach paves the way to real-time applications and to home-monitoring scenarios.


%



\ifCLASSOPTIONcaptionsoff
  \newpage
\fi



\bibliographystyle{IEEEtran}

\end{document}